\documentclass[10pt,twocolumn,letterpaper]{article}

\usepackage{cvpr}
\usepackage{times}
\usepackage{epsfig}
\usepackage{graphicx}
\usepackage{amsmath}
\usepackage{amssymb}
\usepackage{caption}
\usepackage{subcaption}
\usepackage{nopageno}

\usepackage[breaklinks=true,bookmarks=false]{hyperref}

\cvprfinalcopy 

\def\cvprPaperID{****} 

\begin{document}

\title{Image-to-image Neural Network for Addition and Subtraction of a Pair of Not Very Large Numbers}

\author{Vladimir ``vlivashkin'' Ivashkin\\
    Yandex\\
    Moscow, Russia\\
{\tt\small vlivashkin@yandex-team.ru}
}

\twocolumn[{%
\renewcommand\twocolumn[1][]{#1}%
\maketitle
  \centering
  \vspace{-.42in}
  \includegraphics[width=.99\linewidth]{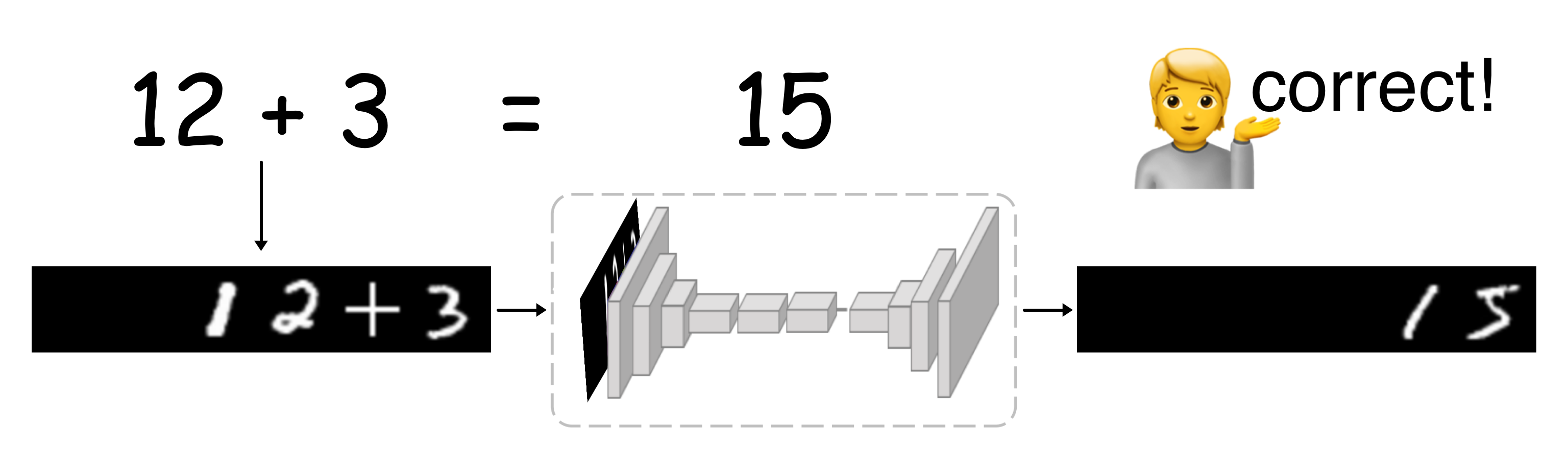}
  \vspace{-.1in}
  \captionof{figure}{We present an image-to-image calculator. First of all, we render an image of a mathematical expression. Then, we feed it to a neural network and get an image of an answer. Finally, we celebrate, but only if the answer is correct.}
  \vspace{.2in}
  \label{fig:teaser}
}]

\maketitle

\begin{abstract}
   Looking back at the history of calculators, one can see that they become less functional and more computationally expensive over time. A modern calculator runs on a personal computer and is drawn at 60 fps only to help us click a few digits with a mouse pointer. A search engine is often used as a calculator, which means that nowadays we need the Internet just to add two numbers. In this paper, we propose to go further and train a convolutional neural network that takes an image of a simple mathematical expression and generates an image of an answer. This neural calculator works only with pairs of double-digit numbers and supports only addition and subtraction. Also, sometimes it makes mistakes. We promise that the proposed calculator is a small step for man, but one giant leap for mankind.
\end{abstract}

\section{Introduction}

Generative Adversarial Networks \cite{goodfellow2014generative} (GANs) are very successfully applied in various computer vision applications, including cats~\cite{this_cat_does_not_exist} and anime generation~\cite{selfie_to_anime}. Still there is not much evidence that they are also good at math.

We follow the history of calculators and present an end-to-end image-to-image neural network calculator, trained with GAN loss. The architecture of this calculator is illustrated in Fig.~\ref{fig:teaser}.

We create such neural calculator which supports addition and subtraction of double-digit numbers.
The demo can be found at \url{https://yandex.ru/lab/calculator?lang=en}.

\section{Related work}
\begin{figure*}
  \centering
  \includegraphics[width=.99\linewidth]{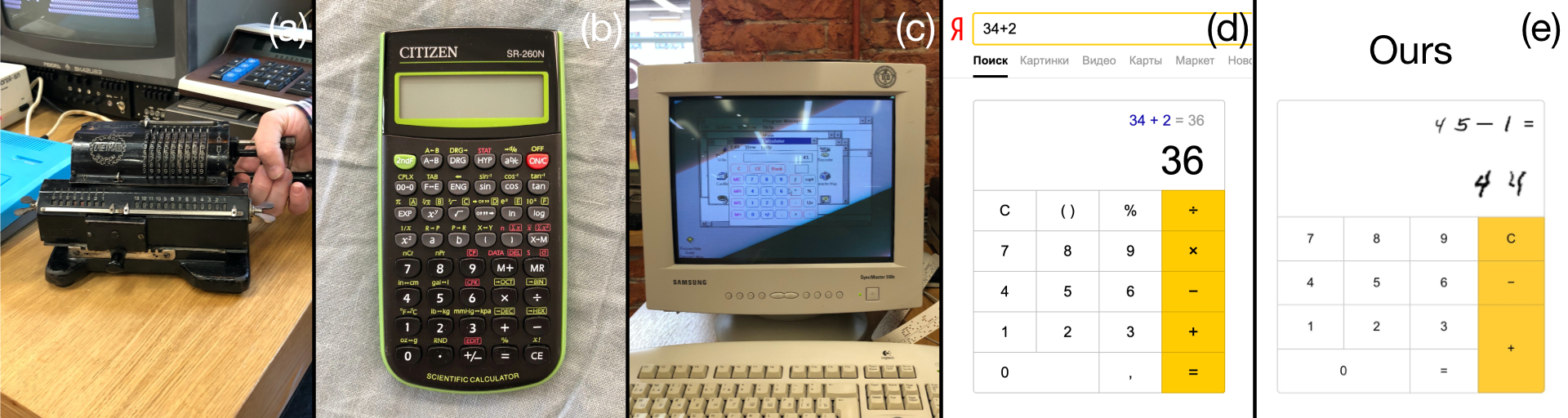}
  \caption{Short history of calculators. (a) a mechanical calculator from 1920s, (b) an electronic pocket calculator from 1980s, (c)  Windows 3.x calculator, (d) search engine calculator, and finally, (e) our solution.}
  \label{fig:short_hist}
\end{figure*}

Calculators always excited humans. The necessity to add and subtract small (and sometimes large) numbers went with the development of human civilization. There is a lot of previous research on this topic, summarized in Fig.~\ref{fig:short_hist}. Let us skip the part with counting on fingers and tally marks and move straight to the Industrial Age.

This mechanical beast from 1920s (see Fig. \ref{fig:short_hist}a) supports addition and subtraction of two nine-digit numbers. In return it needs only a little attention and some twists of the handle. Multiplication and division are also on board but during a ten minute examination we could not figure out how to do it.

The invention of electronic tubes, transistors and microcircuits pushed the development of electronic calculators. The multifunctional battery-powered calculator (e.g. Fig.~\ref{fig:short_hist}b) became the pinnacle of human creation in a physical world. It combines unsurpassed efficiency, usability and functionality. The idea that the epoch of electronic pocket calculators was the best time of human civilization is confirmed by many people and agents. A. Smith~\cite{agent_smith_monologue} said: ``Which is why the Matrix was redesigned to this, the peak of your civilization. I say ``your civilization'' because as soon as we started thinking for you, it really became our civilization which is, of course what this is all about.''

Anyway, then something went wrong: mankind came up with computers. First they operated with punch cards, then with a console, and finally with a graphic interface.
A heavy-duty (relative to the pocket calculator) computer stores an operating system in random-access memory and runs it in an endless cycle, its video card draws 60 frames per second, and all this just to draw a calculator. Monitor shines with pixels instead of using sunlight. The example of such madness is shown in Fig.~\ref{fig:short_hist}c.
Here the functionality of the calculator is simplified, but energy consumption is hundred times increased.

Did we humans stumble in calculator design? Maybe. Did we find the right way? To the best of our knowledge, no. Modern calculators are either an application on some device or even a webpage. Mathematical expressions are among frequent queries in search engines (Fig.~\ref{fig:short_hist}d). In addition to increased capacities and electricity consumption, this method demands the Internet connection (which a very complicated thing) just to add two numbers.

To summarize this survey, calculators are getting slower and simpler in functions. Our calculator (Fig.~\ref{fig:short_hist}e) is a logical extension of previous work on this topic.

\section{Method}
We propose an image-to-image neural network to perform mathematical operations. As there is no suitable dataset for training our model in the literature, we collect our own.

We find that it is possible to create a paired dataset of mathematical expressions, e.g., ``$5+2$'', and corresponding answers, e.g., ``$7$''. Calculators of previous generations are used to collect the data. For each pair of expressions and answers, we generate a pair of images using random MNIST~\cite{lecun-mnisthandwrittendigit-2010} digits of corresponding class.

We choose hourglass UNet~\cite{ronneberger2015u}~-like architecture for our network. The main difference is that we remove all skip-connections and add several linear layers in the bottleneck. It makes the model no longer look like UNet, though. But it helps to prevent network from using parts of an input picture in the output.

Unfortunately, this setup does not allow to train a network just with \textit{L1 loss}. Due to the fact that answer images are built from random MNIST digits, the network converges to generating smooth answers resembling averaged MNIST digits. To encourage the network to produce different lettering, we propose to apply both \textit{GAN-loss}~\cite{goodfellow2014generative} and \textit{perceptual loss}~\cite{johnson2016perceptual}. For perceptual loss we use separate VGG~\cite{simonyan2014very}~-like network trained to recognize MNIST digits.

Calculator operation diagram is shown in Fig.~\ref{fig:teaser}. Neural network takes a rendered expression and returns an image of the result in a form interpretable for humans.

\section{Results}
Using the procedure described above, we successfully trained our neural calculator.
It inputs two integer numbers between $-99$ and $99$ and is able to perform addition or subtraction. According to our experience, this covers almost all daily needs.

\begin{figure}
  \centering
  \includegraphics[width=.99\linewidth]{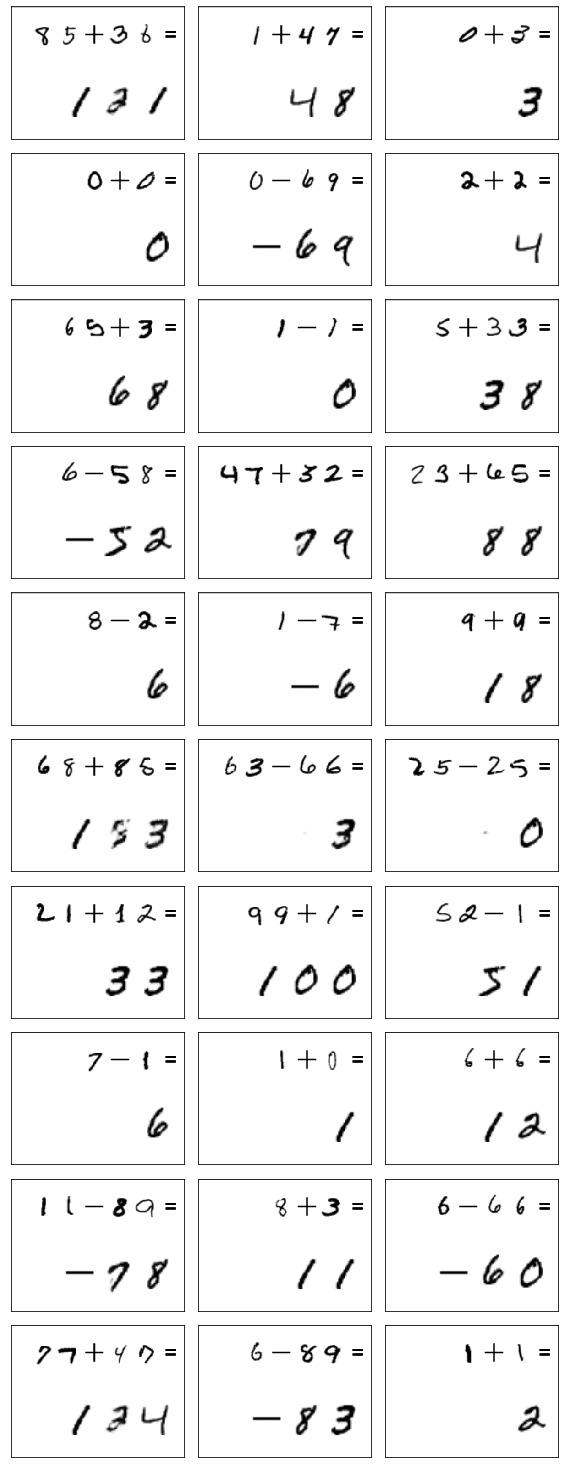}
  \caption{Sample inputs and outputs.}
  \label{fig:sample_outputs}
\end{figure}

Qualitative results of calculations are shown in Fig.~\ref{fig:sample_outputs}. The cherry-picked images show perfect performance of our model. For uncurated results, see our calculator's demo webpage\footnote{\url{https://yandex.ru/lab/calculator?lang=en}}. The comparison of our calculator's performance with the other calculator architectures is presented in Table~\ref{tab:performance}.

\begin{table}
    \centering
    \begin{tabular}{|p{2.35cm}|p{5.05cm}|}
    \hline
    Method & Quality \\
    \hline\hline
    Most calculators & 100\% of success \\
    Ours & We do not use digit recognition as a part of our solution, since this calculator is intended for humans.\\
    \hline
    \end{tabular}
    \caption{Quantitative results of different calculators' performance.}
    \label{tab:performance}
\end{table}

\section{Discussion}
Since we developed this calculator, we have shown it to many influential people in computer vision. Some of them advised to submit this work to SIGBOVIK. We hope that the readers of these proceedings will appreciate the importance of this work and begin to use a more advanced calculator and enter a new calculation era.

We cannot but note that the neural network managed to learn simple arithmetic only from training on images. It is possible to train a model that first turns an image into numbers, performs the arithmetic and then renders an image of the result. This is \textit{not} how our model works. We do not have explicit arithmetic step in our network, but it is still able to generate correct answers.

It could mean that the neural network has mastered the concept of number. The ability to understand concepts and solve problems for which clear rules are not set is what the current neural networks lack in order to become an AGI.

{\small
\bibliographystyle{ieee_fullname}
\bibliography{egbib}
}

\end{document}